\documentclass[11pt,a4paper]{article}
\usepackage{times,latexsym}
\usepackage{url}
\usepackage[T1]{fontenc}
\usepackage[acceptedWithA]{tacl2021v1}

\usepackage{tacl2021v1}
\usepackage{xspace,mfirstuc,tabulary}

\newif\iftaclinstructions
\taclinstructionsfalse % AUTHORS: do NOT set this to true
\iftaclinstructions
\renewcommand{\confidential}{}
\renewcommand{\anonsubtext}{(No author info supplied here, for consistency with
TACL-submission anonymization requirements)}
\newcommand{\instr}
\fi

\iftaclpubformat % this "if" is set by the choice of options

\else

\fi

\usepackage{amsmath}
\usepackage{amssymb}
\usepackage{bbm}
\usepackage{graphicx}
\usepackage{svg}
\usepackage{multirow}
\usepackage{todonotes}
\usepackage{algorithm}
\usepackage{algpseudocode}
\usepackage{caption}
\usepackage{subcaption}
\usepackage{diagbox}
\usepackage[capitalize]{cleveref}
\usepackage{subcaption}
\usepackage{soul}
\usepackage{color}
\usepackage{array}
\usepackage{arydshln}
\usepackage{colortbl}
\usepackage{xcolor}
\usepackage{float}
\title{Source-Free Domain Adaptation for Question Answering with Masked Self-training}

\author{
  Maxwell J. Yin 
  \\
  Western University
  \\
  \texttt{jyin97@uwo.ca}
  \And
  Boyu Wang 
  \\
  Western University
  \\
  \texttt{bwang@csd.uwo.ca}
  \AND
  Yue Dong
  \\
  University of California Riverside
  \\
  \texttt{yue.dong@ucr.edu}
  \And
  Charles Ling\thanks{* Corresponding author} 
  \\
  Western University
  \\
  \texttt{charles.ling@uwo.ca}
}

\date{}

\begin{document}
\maketitle
\begin{abstract}
Previous unsupervised domain adaptation (UDA) methods for question answering (QA) require access to source domain data while fine-tuning the model for the target domain. Source domain data may, however, contain sensitive information and should be protected. In this study, we investigate a more challenging setting, source-free UDA, in which we have only the pretrained source model and target domain data, without access to source domain data. We propose a novel self-training approach to QA models that integrates a specially designed mask module for domain adaptation. The mask is auto-adjusted to extract key domain knowledge when trained on the source domain. To maintain previously learned domain knowledge, certain mask weights are frozen during adaptation, while other weights are adjusted to mitigate domain shifts with pseudo-labeled samples generated in the target domain. %As part of the self-training process, we generate pseudo-labeled samples in the target domain based on models trained in the source domain. 
Our empirical results on four benchmark datasets suggest that our approach significantly enhances the performance of pretrained QA models on the target domain, and even outperforms models that have access to the source data during adaptation. 
\end{abstract}

\section{Introduction}

Question-answering (QA) systems have achieved impressive performance in recent years with the development of pre-trained language models (PLMs). Despite this, research shows that PLMs are sensitive to domain shifts, when training and evaluation datasets have different distributions \cite{yue-etal-2021-contrastive}. In addition, the performance of modern PLMs depends heavily on human-annotated in-domain data when it comes to specific tasks \cite{devlin2018bert,liu2019roberta}. The collection of high-quality datasets is time-consuming and expensive, resulting in many real-world applications relying mostly on unlabeled data.

A possible solution to these issues is to transfer knowledge from a relevant labeled source domain to an unlabeled target domain, which is referred to as unsupervised domain adaptation (UDA). The key to implementing UDA for QA is to extract common features from source and target contexts that provide sufficient information to determine where the answer is located. Following this idea, previous works \cite{wang-etal-2019-adversarial, cao2020unsupervised} utilize adversarial training to make the features extracted from the source and target indistinguishable from each other. Recently, PLMs have become capable of recalling large amounts of information and performing a wide range of tasks  \cite{vaswani2017attention}. Some recent studies \cite{nishida2019unsupervised,cao2020unsupervised,yue-etal-2021-contrastive} incorporate self-training and multitasking training strategies into large PLMs for the task. By simultaneously performing the QA task on the source domain and the language modeling task on the target domain, it is assumed that PLM learns QA knowledge while dealing with domain differences.

However, for these methods to be effective, they must still have access to the source domain data when the models are adapted to target domains. This assumption is often not realistic in real-life applications, as some datasets contain sensitive information with limited access. In these cases, only models trained on these datasets are available to the public. For example, due to privacy concerns in the clinical domain, the organizers of SemEval 2021 Task 10 \cite{laparra-etal-2021-semeval} released models trained on Mayo Clinic clinical notes for public use, yet the original data remain confidential due to ``complicated data use agreements". 

Our study explores source-free UDA, a setting where users can adapt models developed on private data to their own target domain without access to private source data. Historically, this has been a very difficult setting. Recent advancements in transformer-based PLMs \citep{gururangan-etal-2020-dont,yin2024fast} suggest that these systems possess a remarkable capacity for learning domain-specific knowledge and effectively answering queries. It is therefore feasible to explore how trained models can be adapted to target domains in the absence of original source data.

We propose a novel approach based solely on self-training, called masked domain adaptation for question answering (MDAQA).  A bottleneck module is designed first with a masking mechanism. This bottleneck module is then inserted between the QA encoder and the final prediction layer. When the QA model is trained on the source domain, the mask is learned automatically to pass only values of specific kernels of the bottleneck module to the prediction layer, while the remaining kernels are left unused. When the QA model is adapted to the target domain, weights of links to early mentioned kernels are frozen in order to keep previously learned domain knowledge, while the weights related to remaining kernels are adjusted to mitigate the domain variance. Since the target domain data are unlabeled, we use the QA model trained on the source domain to make predictions on the target domain data. Then we use predictions with high confidence as pseudo labels to further fine-tune our model\footnote{The code for our MDAQA framework is available at \url{https://github.com/maxwellyin/MDAQA}}.
 
We evaluate MDAQA on well-known QA benchmark datasets. Experiment outcomes show that MDAQA significantly improves the performance of pretrained QA models on target domains. MDAQA even outperforms previous UDA methods, which require access to source data during adaptation. To summarize, the proposed method has the following features:

\begin{itemize}
    \item We introduce MDAQA, an innovative self-training-based domain adaptation technique specifically tailored for QA applications. This approach features a masking module, designed to facilitate the learning and retention of domain-specific knowledge by QA models. Furthermore, it mitigates the detrimental effects associated with domain shifts during the adaptation to target domains.
    \item The masking module we have engineered is lightweight and modular, allowing for seamless integration into a broad array of existing QA models. This design feature enhances the applicability of the method and ensures ease of use in diverse scenarios.
    \item Unlike previous approaches, MDAQA eliminates the need for direct comparison between source and target domain data during the adaptation process. This characteristic renders our method especially advantageous when handling source data that may contain sensitive information.
\end{itemize}

\section{Related Work}

In this section, we briefly review two lines of related work. Firstly, existing UDA for QA works. Secondly, the recently proposed source-free UDA works.

\subsection{Unsupervised Domain Adaptation of Question Answering}

QA systems are designed to generate answers from text, with a significant focus on extractive QA — identifying answer spans within unstructured text based on a posed question. This aspect of UDA has been thoroughly explored, beginning with the pioneering work by \citet{wang-etal-2019-adversarial}, which brought domain adversarial training into the realm of QA, extending the foundational work by \citet{ganin2016domain}. Subsequent research has built upon this by exploring various strategies to enhance domain adaptation. For example, \citet{nishida2019unsupervised} introduced a multitask training approach, \citet{cao2020unsupervised} combined domain adversarial training with self-training techniques, and \citet{yue-etal-2021-contrastive} applied contrastive learning to distinguish between answer and context token embeddings. In parallel, input-level masking has emerged as another innovative pathway, where pivot features—elements critical to the task and common to both domains—are masked during pre-training, leading to the creation of domain-agnostic representations \citep{blitzer-etal-2006-domain,ziser-reichart-2018-pivot, ben-david-etal-2020-perl, lekhtman-etal-2021-dilbert}. Despite the progress made, a common limitation of these methods is the necessity for concurrent access to data from both source and target domains.

\subsection{Source-free Unsupervised Domain Adaptation}

Source-free UDA studies the more restricted case of UDA where the source domain data is not available when we adapt the model trained on the source domain to the target domain. Most previous research has focused on classification problems. Source-free UDA is first proposed by \citet{liang2020we}. They assume the classification outputs for both source and target domain data should be similar to each other. Therefore, they leverage the idea of information maximization \cite{hu2017learning} and fine-tune their encoder to maximize the mutual information between intermediate feature representations and outputs of the classifier. 3C-GAN \cite{li2020model} is another representative work. Given the trained model and target domain data, they use generative adversarial networks \cite{goodfellow2020generative} to produce target-style training samples. \citet{huang2021model} introduce historical contrastive learning into this research field, which contrasts the embeddings generated by the currently adapted model and the historical models. 

Some researchers have introduced Source-free UDA into the study of natural language processing recently. \citet{zhang-etal-2021-matching} combine knowledge distillation and source-free UDA together for sentence classification. Specifically, they modify the joint maximum mean discrepancy \citep{long2017deep} to match the joint distributions between a trained source model and target domain samples while performing knowledge distillation. \citep{yi2023source} explore the intersection of source-free UDA and learning with noisy labels, addressing the noise in pseudo-labels generated by source models due to domain shifts. \citet{zeng2022domain,wang2023dynamically} extend Source-free UDA to more complicated setting.

Nevertheless, the primary focus of these studies lies in classical classification problems or tasks that can be adapted into such formats. As a result, the majority of these methods are developed with the premise that each specific sample correlates to a distinct class label. In stark contrast, QA tasks require a more nuanced approach.For each input sample in QA tasks, a sequential classification is required for the series of tokens it comprises. This process involves determining the role of each token—identifying it as either a start token, an end token, or a regular token within the sequence. This classification underscores the importance of context, as the significance of each token is defined not in isolation but in relation to the entire sequence of tokens. This process is intrinsically context-dependent, making the classification of isolated tokens substantially less meaningful. Due to this fundamental difference, most existing source-free UDA techniques are not directly applicable to QA tasks.

\section{Problem Definition}

For extractive QA, given a supporting context $p=(p_1,p_2,...,p_{L_1})$ with $L_1$ tokens and a query $q=(q_1,q_2,...,q_{L_2})$ with $L_2$ tokens, the answer $a=(p_{a^s},p_{a^s+1},...,p_{a^e})$ is a text piece in the original passage $p$. This goal of extractive QA is to find out the correct answer span $(a^s,a^e),0<a^s<a^e<L_1$, where $a^s,a^e$ represent the starting and ending indexes of the answer in passage $p$.

Source-free UDA for QA can then be formally defined as follows: There exists a source domain $\mathcal{D}_S$, which possesses labeled data that is only accessible during the initial training of the source model. A target domain $\mathcal{D}_T$ exists, which has freely accessible but unlabeled data. In the source domain $\mathcal{D}_S$, we utilize $n$ labeled samples $\{x_i,y_i\}^n_{i=1}$, where the input $x_i=(p_i,q_i)$ and label $y_i=(a^s_i,a^e_i)$. We then have $n^\prime$ unlabeled samples $\{x^\prime_j\}^{n^\prime}_{j=1}$ from the target domain $\mathcal{D}_T$, adhering to the same QA task as aforementioned. We postulate that the data distribution in $\mathcal{D}_S$ and $\mathcal{D}_T$ are not identical. The initial step is to train a model on $\mathcal{D}_S$, after which, access to $\mathcal{D}_S$ ceases. The core of the process, however, lies in the subsequent adaptation of the pre-trained model to $\mathcal{D}_T$. This transition is designed to maximize performance on the target domain by proficiently mitigating the domain shift between $\mathcal{D}_S$ and $\mathcal{D}_T$.

\section{Method}

In this section, we introduce our proposed approach, Masked Domain Adaptation for Question Answering (MDAQA). Our primary motivation is to leverage the domain knowledge acquired from the source domain and adapt it efficiently for the target domain within the context of QA. The ideal model should be capable of discerning and retaining vital information while adapting to the differences between the source and target domains, thus alleviating domain shift. To achieve this, we propose MDAQA, which consists of a feature extraction base model and a specially designed mask module. The mask module plays a pivotal role by controlling the activation and usage of specific features or kernels in the prediction layer, thereby focusing the model's learning process on pertinent domain knowledge.

Following this, we delve into the architecture of the base model and the specialized mask module. We also detail the training processes implemented for both the source and target domains.

\subsection{Base Model}

Our feature extractor is the pre-trained language model ROBERTA \cite{liu2019roberta}. We fine-tune it for QA following the setup proposed by \cite{nishida2019unsupervised}. On top of this feature extractor, we add a linear layer for extractive question answers, whose output dimension is two. We then add a softmax layer to map these dimensions to probabilities, representing the probability that a token is the start or end of an answer span. Given a pair of $(p^i,q^i)$, the input sequence is [<s>; $q^i$; </s></s>; $p^i$; </s>]. We then attribute a score to each possible answer span by taking the product of corresponding probabilities. The answer span with the highest score is selected as the predicted answer. For convenience, we denote the feature extractor as $g$ and the final prediction layer as $h$. Therefore, the final output for this base model can be denoted as $\hat{y}(x)=(h\circ g)(x)$.

\subsection{Mask Module}

\begin{figure}
\begin{center}
\includegraphics[width=\columnwidth]{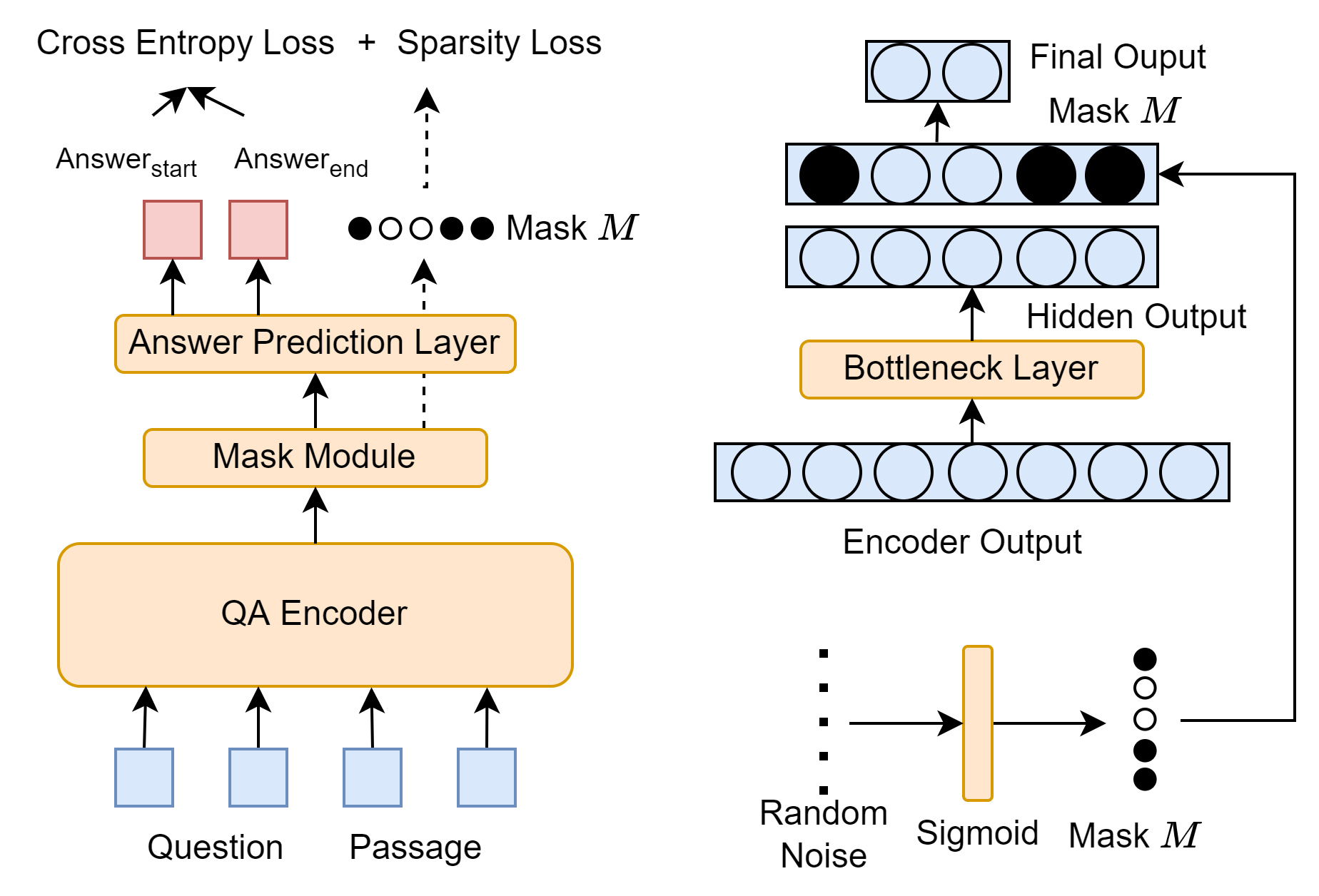}
\end{center}
\caption{Illustration of the proposed MDAQA architecture. The left part is the overall architecture, we insert a mask module between the QA encoder and the answer prediction layer. The QA encoder can be any common language model. Here we use ROBERTA. The right part is the detailed architecture of the mask module.} \label{model}
\end{figure}

The mask module, illustrated in Figure \ref{model}, consists of a linear bottleneck layer $f$ and a mask vector $M$. The mask vector $M$ is designed to learn and select valuable features by taking on close-to-binary values during training. Specifically, 
\begin{equation}
    M=\sigma (k\cdot N)
\end{equation}
where $\sigma$ is the sigmoid function, $N$ is another learnable vector with initial values sampled from a uniform distribution on the interval $[-0.5,0.5]$, and $k$ is a large hyperparameter to make sure the output of $\sigma$ is nearly binary.
The mask module is inserted between the feature extractor $g$ and the prediction layer $h$. The linear bottleneck layer is designed to project the output of $g$ to a lower dimensional feature space. It also makes the mask module separate from $g$ and $h$, allowing for more flexibility in combining the mask modules into various QA models. Thus, the output feature of $f$ can be denoted as $(f\circ g)(x)$. Then the mask vector $M$ is applied to it, thus only certain channels of the feature embedding are activated and passed to $h$. Therefore, the final output is changed to:
\begin{equation}
    \hat{y}(x)= h(M \odot (f\circ g)(x))
\end{equation}

where \(M\) is broadcasted over the matrix during the element-wise multiplication, denoted by \(\odot\).

\subsection{Training on Source Domain}

\begin{figure}
\begin{center}
\includegraphics[width=\columnwidth]{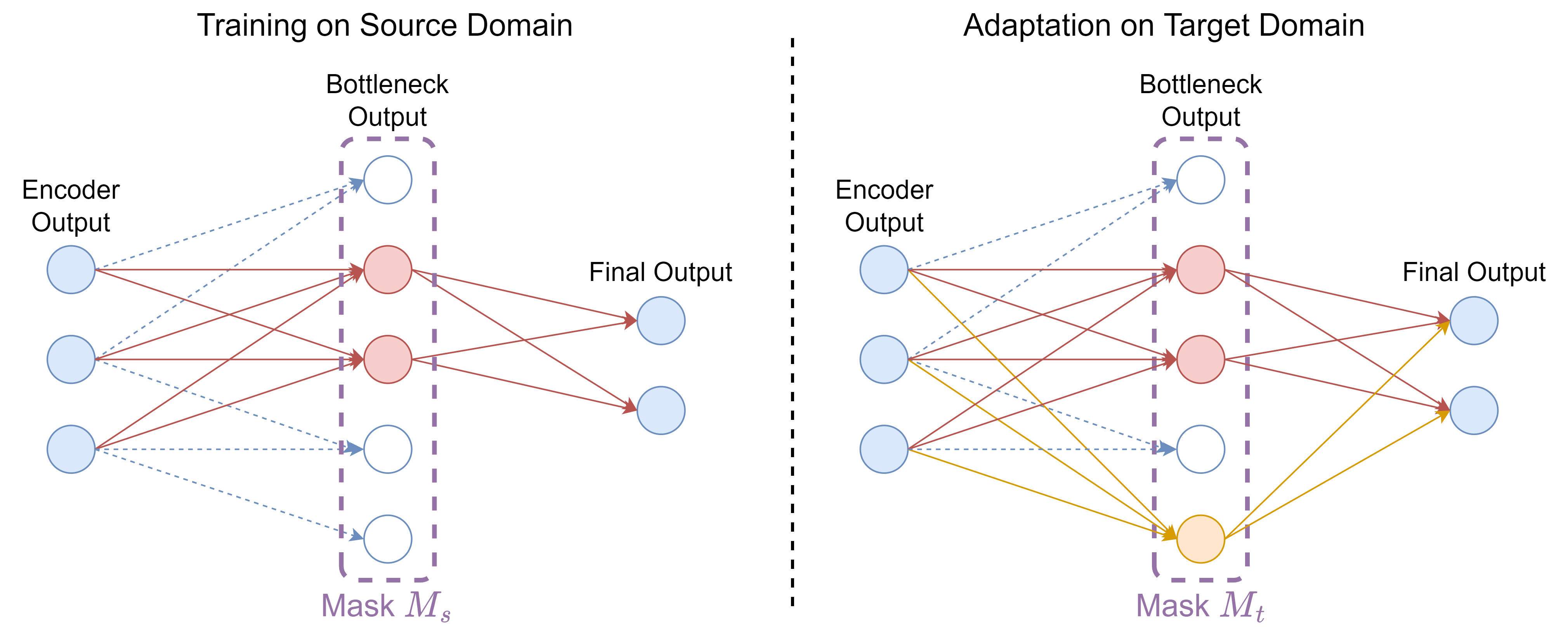}
\end{center}
\caption{A concise illustration of the weight adjustment associated with the mask module during source training and adaptation. All weights are free to change when the model is trained on the source domain. The red circles are kernels with mask values close to 1; thus, their output is passed to the answer prediction layer. During adaptation, the links marked in red experience minimal weight adjustments, in contrast to the remaining weights, which continue to adjust dynamically. If the mask value of some kernel changes from 0 to 1, denoted by the yellow circle, the output of that kernel will also be passed to the final layer.} \label{mask}
\end{figure}

During the training process on the source domain, we minimize the cross-entropy loss between the final outputs $\hat{y}$ and labels $y$, and add a sparsity loss to the mask vector $M$ to preserve only the most significant features. The loss function is defined as:
\begin{equation}
    \mathcal{L}=\mathcal{L}_{ce}(\hat{y},y)+\lambda \cdot sum(M)/b
    \label{eq:loss}
\end{equation}
where $sum(M)$ is the sum of all values in $M$ and $\lambda$ is a hyperparameter.

\subsection{Training on Target Domain}

\subsubsection{Mask adaptation}

As depicted in Figure \ref{mask}, after training on the source, the mask vector $M$ should learn to activate only important kernels of the bottleneck layer output $(f\circ g)(x)$, which contain domain knowledge as well as QA knowledge, and the rest of the kernels are masked. We denote its current state as $M_c$.

As the model adapts to the target domain, we expect it to retain previously learned knowledge while learning to mitigate the effects of domain shift. Therefore, we keep the weights of links directly connected to previously activated kernels fixed, while further adjusting the weights of the remaining links to improve performance on the target domain. Formally, assume $W_f\in \mathbb{R}^{b\times a}$ is the weight of the bottleneck layer $f$, $W_h\in \mathbb{R}^{2\times b}$ is the weight of the answer prediction layer $h$. We use the previously learned $M_c$ to regularize the gradients flowing through the mask module during backpropagation:
\begin{equation}
    W_f \leftarrow W_f - (\bar{M}_c \mathbbm{1}^T_{a})\odot \frac{\partial \mathcal{L}}{\partial W_f}
\end{equation}
\begin{equation}
    W_h \leftarrow W_h - \frac{\partial \mathcal{L}}{\partial W_h}\odot (\mathbbm{1}_2 \bar{M}_c^T)
\end{equation}
where $\bar{M}_c=1-M_c$, $\mathbbm{1}_d$ is an all ones vector of dimensionality $d$. Since the values in $M_c$ are all close to 0 or 1, this ensures that the key features that are related to previously learned knowledge are only minimally changed, while newly activated kernels focus on handling the domain shift. For better flexibility and generalization ability, the QA encoder is also fine-tuned during the adaptation process. However, we set changes to it to a much smaller adjustment range than the mask module.

\subsubsection{Self-training}

Since there is no labeled sample in the target domain, we leverage a self-training strategy \cite{yarowsky-1995-unsupervised,mcclosky-etal-2006-effective} to generate pseudo labels. Specifically, we note that the source domain $D_s$ and the target domain $D_t$ may share some common characteristics even though they have different distributions. Therefore, given a model $F(x)=h(M_c\odot (f\circ g)(x))$ learned on the source domain $\mathcal{D}_s$, and unlabeled samples $\{x'_j\}^{n'}_{j=1}$ from the target domain $\mathcal{D}_t$, some predicted answers from $\{\hat{y}'_j\ |\ \hat{y}'_j=F(x'_j), j=1,...,n'\}$ should be similar to the correct answer spans for the corresponding samples in $\mathcal{D}_t$. Therefore, these predicted answers can be used as pseudo labels for corresponding samples and further fine-tune the trained model $F$.

As mentioned earlier, our model will assign a likelihood score to each predicted answer. To avoid significant error propagation, we select predictions of high confidence as pseudo labels. Assume the score is $s_j$ for $\hat{y}'_j$, we can obtain a subset of target samples with pseudo labels: $\{(x'_j,\hat{y}'_j)\ |\ s_j>\alpha, j=1,...,n'\}$, where the threshold $\alpha$ is a hyperparameter. We use this subset as the training dataset to fine-tune the trained model $F$.

When we fine-tune the model $F$ on the target domain $\mathcal{D}_t$, we use the same loss function as depicted in Eq. \ref{eq:loss}.  We repeat this process of pseudo-label dataset generation and model fine-tuning multiple times. We keep $\alpha$ the same value in all iterations. As the number of iteration increases, the model $F$ should gradually learn new domain knowledge of $\mathcal{D}_t$ and thus become more confident about its own prediction. Therefore, the size of the generated dataset should increase accordingly with the number of iterations. The full training procedure is illustrated in Algorithm \ref{alg:1}.

\begin{algorithm}[ht]
\caption{MDAQA Training and Adaptation}
\label{alg:1}
\begin{algorithmic}
\State \textbf{Input:} labeled samples $\{x_i,y_i\}^n_{i=1}$ in the source domain $\mathcal{D}_s$. Unlabeled samples $\{x'_j\}^{n'}_{j=1}$ in the target domain $\mathcal{D}_t$. Pretrained QA base model $y(x)=(h\circ g)(x)$.  The source domain training epoch number is $N_s$. Domain adaptation training epoch number is $N_t$.
\State \textbf{Output:} Optimal model in the target domain.
\end{algorithmic}
\begin{algorithmic}[1]
\State Insert the proposed mask module into the pretrained model and get the MDAQA model $F$: $y(x)= h(M\odot (f\circ g)(x))$
\For{$i\leftarrow 1$ to $N_s$}
    \State Train $F$ with mini-batch from $\{x_i,y_i\}^n_{i=1}$.
\EndFor
\For{$i\leftarrow 1$ to $N_t$}
    \State Pseudo labeled set $S^P \leftarrow \emptyset$
    \For{$j\leftarrow 1$ to $n'$}
        \State Use $F$ to predict the label $\hat{y}'_j$ for $x'_j$
        \State Obtain probability score $s_j$
    \EndFor
    \If{$s_j>\alpha$}
        \State Put $\{x'_j,\hat{y}'_j\}$ into $S^P$
    \EndIf
    \For{mini-batch $\mathcal{B}$ in $S^P$}
        \State Train $F$ with mini-batch $\mathcal{B}$
    \EndFor
\EndFor

\end{algorithmic}
\end{algorithm}

\section{Experiment Setup}

In this section, we introduce our experimental setup. We first introduce our chosen dataset and the necessary preparations. We then introduce baselines for comparison. Finally, we show the details of the model implementation.

\subsection{Datasets}

Following previous work \cite{nishida2019unsupervised,yue-etal-2021-contrastive,yue-etal-2022-synthetic}, We use datasets in the MRQA 2019 Shared Task \cite{fisch-etal-2019-mrqa}. In our experiments, we use SQuAD \cite{rajpurkar-etal-2016-squad} as our source domain dataset. For the target domain, we consider HotpotQA \cite{yang-etal-2018-hotpotqa}, Natural Questions (NQ) \cite{kwiatkowski2019natural}, NewsQA \cite{trischler-etal-2017-newsqa}, and BioASQ \cite{tsatsaronis2015overview} as they are commonly used benchmark datasets.  For target domain datasets, only unlabeled samples are accessible for adaptation. In order to simulate a low-resource setting that is commonly used in UDA research, we randomly choose $n=1,000$ unlabeled samples (about 1\%-1.5\% of original training sets) from each target domain training dataset for domain adaptation. We also discuss the impact of $n$ in Section \ref{sec:data_size}.

To more accurately represent the real-world application of Source free UDA, we conduct an extended experiment to assess the UDA performance of our method and baselines in adapting the model from SQuAD to a clinical dataset of limited size. This dataset, sourced from MIMIC-III \citep{yue2021cliniqg4qa,Yue2021Annotated,Goldberger2000}, has been manually modified to conform to the SQuAD format.

\subsection{Baselines}

In this study, we juxtapose MDAQA with the subsequent advanced benchmarks: AdaMRC \cite{wang-etal-2019-adversarial}, UDARC \cite{nishida2019unsupervised}, and CAQA \cite{yue-etal-2021-contrastive}. Given the divergence in QA backbone models utilized in these works, we undertake a reimplementation using the ROBERTA framework \cite{liu2019roberta}, aligning with our employed model, to ensure an equitable comparison.

Furthermore, we introduce a ``No Adaptation" baseline, where the model is trained exclusively on the source domain dataset and subsequently assessed on the target domain datasets. This strategy aims to gauge the foundational performance in the absence of any prior knowledge pertaining to the target domain.

In our extensive evaluation, we also introduce a ``Supervised Learning'' baseline. In this configuration, the model is granted unfettered access to the entirety of the labeled target data. This baseline serves as a benchmark, setting a potential performance ceiling. It aids in highlighting the relative efficacy of our UDA approach compared to scenarios with complete target domain knowledge.
    
To provide a comprehensive analysis of our proposed methodology, we conduct several ablation studies to examine its different components. The variant labeled ``No Mask'' allows us to understand the performance metrics of MDAQA without the inclusion of the mask module. On the other hand, the ``Random Mask'' strategy retains the mask module but initializes its weights randomly and keeps them constant. This serves to highlight the importance of having adaptable mask weights in our model.

To further explore the effects of the training process on domain adaptation, a ``One Round'' baseline is incorporated. This baseline examines the performance of the model when self-training is conducted in a single iteration. This evaluation aims to determine the significance of conducting multiple iterations of training and to understand the impact of a singular training round on domain adaptation.

Building upon the baselines, we further extend our analysis by comparing our mask module with traditional adapter modules for UDA of QA. This comparative study is pivotal for demonstrating the robustness and innovation of our approach.

\subsection{Implementation Details}

All models, including baseline methods and our proposed approach, are built upon the foundational model \texttt{roberta-base} from the HuggingFace transformers library \cite{wolf-etal-2020-transformers}. To ensure a fair comparison, we adhered strictly to default parameter settings of the library, except where explicitly stated. We set the hyperparameters as \( k=100 \), \( \lambda=0.75 \), and \( \alpha \) within the range of 0.5 to 0.7. The selection of \( \alpha \) is based on the model selection strategy \citep{nguyen2020leep,yang2023can}, and the self-training process is executed for 5 rounds. The dimensionality of the bottleneck layer is 256. For optimization, we employ the AdamW optimizer. When the model undergoes training on the source domain, the learning rate for the mask module is 1e-3 and for the remaining modules is 2e-5. During adaptation to the target domain, the learning rate for the mask module is adjusted to 5e-4, while for other modules it remains at 2e-5. Given the absence of test sets in the MRQA datasets, we partition the original dev sets into new dev and test subsets at a 1:1 ratio. For training in the source domain, we utilize all available training data, while for domain adaptation, only \( n=1000 \) unlabeled samples are used, as mentioned earlier.

For the ``No Masking'' ablation study, we trained the model without the mask module, using the same learning rates as the full model.  In contrast, the ``Random Masking'' scenario employed the mask module, but with its weights initialized to random values sampled from a uniform distribution, which remained fixed during training. For the adapter-based approach, we follow the architecture proposed by \citet{houlsby2019parameter}, with the learning rate for the adapter set to match that of the mask module. Across all ablation studies, we ensured that all other parameters were kept consistent, thereby distinctly attributing any variations in model performance to the effects of the masking and adapter mechanisms.

\begin{table*}[ht]
    \centering
    \small
    \begin{tabular}{l|cc|cc|cc|cc|cc}
    \hline
    \multirow{2}{*}{Methods} & \multicolumn{2}{c|}{HotpotQA} & \multicolumn{2}{c|}{NQ} & \multicolumn{2}{c|}{NewsQA} & \multicolumn{2}{c|}{BioASQ} & \multicolumn{2}{c}{MIMIC-III} \\
    \cline{2-11}
    & EM & F1 & EM & F1 & EM & F1 & EM & F1 & EM & F1 \\
    \hline
    No Adaptation & 47.94 & 65.71 & 46.39 & 60.17 & 41.12 & 57.20 & 49.77 & 61.69 & 14.57 & 47.76 \\
    UDARC & 48.51 & 66.40 & 46.93 & 60.49 & 41.18 & 57.26 & 50.11 & 62.51 & 15.63 & 50.12 \\
    AdaMRC & 50.13 & 67.47 & 48.40 & 61.11 & 41.35 & 57.31 & 51.85 & 63.25 & 16.49 & 50.36 \\
    CAQA & 51.28 & 68.85 & 51.10 & 64.01 & 44.29 & 59.02 & 52.64 & 63.09 & 18.12 & 51.78 \\
    \hline
    No Mask & 50.73 & 66.81 & 50.03 & 62.92 & 43.18 & 57.98 & 51.07 & 61.95 & 16.04 & 48.48 \\
    Random Mask & 50.60 & 66.52 & 49.35 & 62.21 & 43.03 & 56.72 & 50.12 & 61.19 & 15.79 & 48.29 \\
    One Round & 52.71 & 68.66 & 52.65 & 65.13 & 44.03 & 60.29 & 53.11 & 63.04 & 17.02 & 51.13 \\
    MDAQA & \textbf{53.81} & \textbf{70.34} & \textbf{53.43} & \textbf{66.26} & \textbf{44.89} & \textbf{60.50} & \textbf{54.12} & \textbf{64.07} & \textbf{19.03} & \textbf{53.94} \\
    \hline
    \end{tabular}
    \caption{Main results on comparing question-answering performance while performing domain adaptation from SQuAD to MRQA datasets and MIMIC-III. EM denotes the exact match.}
    \label{tab:main-results}
\end{table*}

\section{Results}

\begin{figure*}
    \begin{center}
    \includegraphics[width=\textwidth]{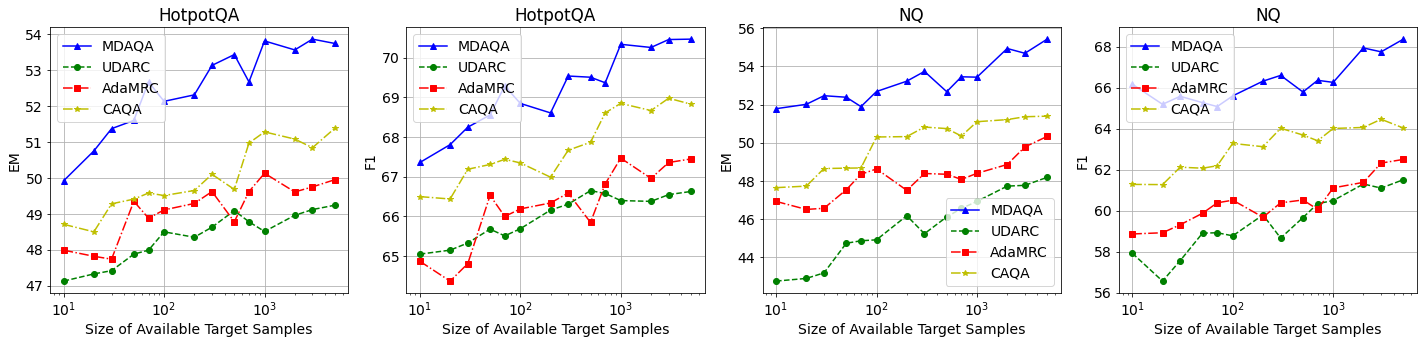}
    \end{center}
    \caption{Impact of the number of available unlabeled target domain samples. We show the performance of MDAQA and baselines measured by EM and F1 when adapted to HotpotQA or NQ. We use logarithmic scaling for the x-axis because the performance change of the algorithms is more pronounced at low data volumes}
    \label{fig:data_size_fig}
    \end{figure*}

\begin{figure}[ht]
    \begin{center}
    \includegraphics[width=\columnwidth]{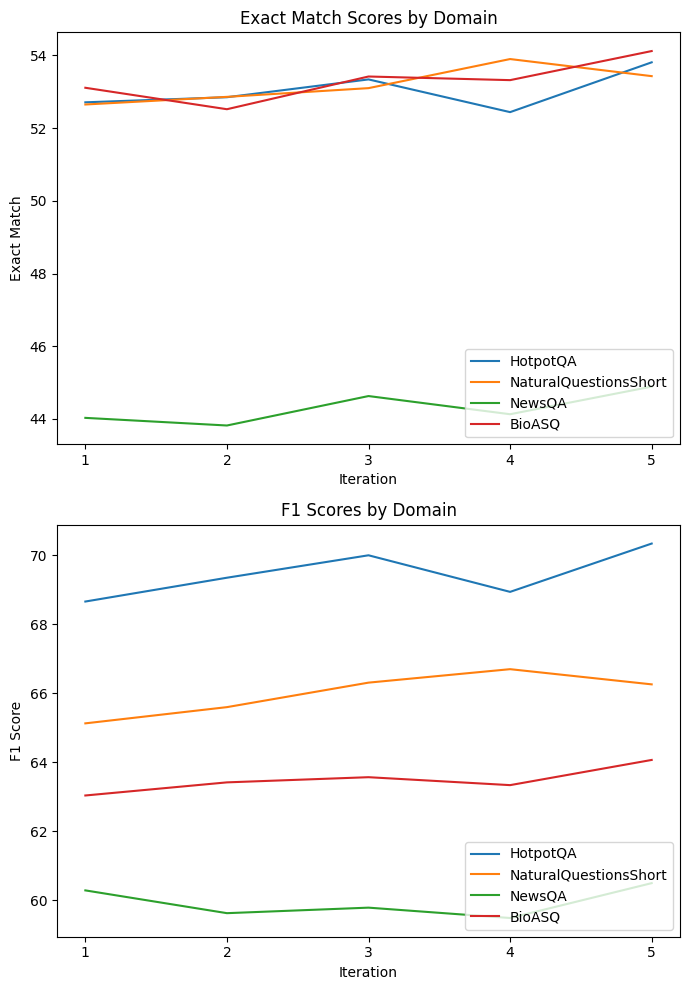}
    \end{center}
    \caption{{Performance Evolution Across Domains Over Five Rounds of self-training}}
    \label{fig:sf-iteration}
\end{figure}

In this section, we present our experimental results and analyze them. We first present the experimental performance of MDAQA compared to baselines on benchmark datasets, as well as the ablation study. We then analyze the impact of the number of available target domain samples. Finally, we analyze the effect of the threshold $\alpha$.

\subsection{Performance on Target Domains}

Table \ref{tab:main-results} reports the main results. As indicated in the table, the No Adaptation baseline is consistently outperformed by our proposed MDAQA framework. Compared to this baseline, MDAQA leads to a performance improvement in EM of at least 3.77\% and can go as high as 7.04\%, and an improvement in F1 of at least 3.30\% and up to 6.09\%. This verifies that the performance of pretrained QA models suffers from domain shift when used on target domains that have a different distribution than the source domain from which they are trained, and that UDA methods have 
good application value in such circumstances. 

MDAQA also outperforms all remaining baselines by a clear margin. CAQA yields the best performance among all the baselines. Compared with it, our proposed MDAQA leads to a performance improvement in EM of at least 0.60\% and can go as high as 2.43\%, and an improvement in F1 of at least 0.98\% and up to 2.25\%. This suggests that MDAQA is better at handling the effects of domain shift than previous methods. Since MDAQA does not need to access source data when doing domain adaptation, it also has wider applications.

The performance of UDA methods varies according to different target domain datasets. As depicted in Table \ref{tab:main-results}, the NewsQA seems to be the hardest one for domain adaptation. The No Adaptation baseline has an EM of 41.12 and an F1 of 57.20, significantly lower than its performance on other datasets. The UDA methods also yield minimal improvement on this dataset. Compared with the No adaptation baseline, our proposed MDAQA only improves EM by 3.77\% and F1 by 3.30\%. It is also only slightly better than CAQA. This may be because NewsQA is the only dataset based on news articles, while the articles in the rest of the datasets are more academic. On the other hand, the UDA methods seem to work best on NQ dataset. Compared with the No adaptation baseline, our proposed MDAQA improves EM by 7.04\% and F1 by 6.09\%. This may be because that NQ is also an open-domain QA dataset based on Wikipedia articles, the same as SQuAD.

Notably, when adapting from SQuAD to MIMIC-III, a considerable performance decrease is observed for all methods compared to their performance on MRQA datasets. This decline can be attributed to the substantial domain gap between the source and target domains, with the MIMIC-III dataset potentially incorporating more noise due to differences in data processing levels. Despite these challenges, MDAQA still achieves the highest scores, reinforcing its effectiveness in diverse and challenging real-world scenarios.

\subsection{Ablation Study}
    
This ablation study is carried out to systematically clarify the importance of the mask module in our proposed framework. The data shown in Table \ref{tab:main-results} confirm that MDAQA consistently outperforms its baselines. A comparison between MDAQA and its "No Mask" variant shows an improvement in the EM metric that varies from 0.71\% to 3.40\%, and in the F1 metric that ranges from 1.52\% to 3.34\%. These findings highlight the critical role of the mask module in preserving domain-specific knowledge and effectively handling domain shifts.
    
Furthermore, when examining the "Random Mask" row in Table \ref{tab:main-results}, it is worth noting that MDAQA consistently outperforms this condition as well. Specifically, the EM and F1 metrics are markedly higher in MDAQA compared to the "Random Mask" configuration, lending credence to our claim that the mask module is not just a random operation but is systematically improving performance.

{
    The "One Round" approach in Table \ref{tab:main-results}, which involves a single iteration of self-training, presents competitive performance. Furthermore, the iterative nature of the self-training process in the MDAQA model further enhances the results. As depicted in Figure \ref{fig:sf-iteration}, across the four target domains, there is a general trend of performance improvement with each subsequent round of self-training, culminating in the highest scores at Round 5. This pattern underscores the progressive learning capability of the MDAQA architecture, where each round builds upon the previous, fine-tuning the adaptability and accuracy of our model.

    These observations collectively affirm the efficacy of the iterative self-training and the mask module in the MDAQA architecture, validating their roles in enhancing question-answering performance across diverse domains.
}

\subsection{Comparative Study: Mask Module vs. Adapter}

In this section, we offer a concentrated analysis between our proposed mask module and traditional adapter techniques in the field of question-answering. Specifically, as delineated in Table~\ref{tab:adapter-vs-mdaqa}, the "Adapter" model substitutes our mask module with a conventional adapter to enable a direct comparison with established adapter-centric methods.

As shown in Table~\ref{tab:adapter-vs-mdaqa}, our MDAQA method excels in both EM and F1 scores across multiple datasets, thereby substantially outperforming the adapter model. This edge is principally attributed to the distinct capabilities of our mask module: it not only offers fine-grained domain knowledge retention but is also designed for computational and memory efficiency. Moreover, it is specifically engineered to tackle the unique challenges encountered in QA domain adaptation, particularly in source-free settings.

\begin{table*}[ht]
    \centering
    \small
    \begin{tabular}{l|cc|cc|cc|cc|cc}
    \hline
    \multirow{2}{*}{Methods} & \multicolumn{2}{c|}{HotpotQA} & \multicolumn{2}{c|}{NQ} & \multicolumn{2}{c|}{NewsQA} & \multicolumn{2}{c|}{BioASQ} & \multicolumn{2}{c}{Average} \\
    \cline{2-11}
    & EM & F1 & EM & F1 & EM & F1 & EM & F1 & EM & F1 \\
    \hline
    Adapter & 51.90 & 67.14 & 51.29 & 64.06 & 43.53 & 58.49 & 52.39 & 62.61 & 49.78 & 63.08 \\
    MDAQA & \textbf{53.81} & \textbf{70.34} & \textbf{53.43} & \textbf{66.26} & \textbf{44.89} & \textbf{60.50} & \textbf{54.12} & \textbf{64.07} & \textbf{51.56} & \textbf{65.29} \\
    \hline
    \end{tabular}
    \caption{Focused comparison of Adapter and MDAQA methods on question-answering performance during domain adaptation. EM denotes exact match.}
    \label{tab:adapter-vs-mdaqa}
\end{table*}

\subsection{Mask Module in UDA Compared with Supervised Domain Adaptation}

\begin{table*}[ht]
    \centering
    \small
    \begin{tabular}{l|cc|cc|cc|cc|cc}
    \hline
    \multirow{2}{*}{Methods} & \multicolumn{2}{c|}{HotpotQA} & \multicolumn{2}{c|}{NQ} & \multicolumn{2}{c|}{NewsQA} & \multicolumn{2}{c|}{BioASQ} & \multicolumn{2}{c}{Average} \\
    \cline{2-11}
    & EM & F1 & EM & F1 & EM & F1 & EM & F1 & EM & F1 \\
    \hline
    UDA & 53.81 & 70.34 & 53.43 & 66.26 & 44.89 & 60.50 & 54.12 & 64.07 & 51.56 & 65.29 \\
    -Mask & 50.73 & 66.81 & 50.03 & 62.92 & 43.18 & 57.98 & 51.07 & 61.95 & 48.75 & 62.42 \\
    \hline
    Supervised Learning & 63.58 & 79.69 & 68.68 & 80.02 & 57.19 & 72.49 & 67.51 & 75.53 & 64.24 & 76.93 \\
    -Mask & 63.48 & 79.41 & 67.16 & 79.33 & 56.86 & 71.91 & 66.86 & 75.01 & 63.59 & 76.42 \\
    \hline
    \end{tabular}
    \caption{The performance of MDAQA under UDA and supervised learning scenarios with and without the mask module. The "-Mask" rows indicate performance when the mask module is not included.}
    \label{tab:mdaqa-results-across-scenarios}
\end{table*}
    
To further validate the efficacy of the mask module, experiments were extended to transfer learning scenarios, with the results summarized in Table \ref{tab:mdaqa-results-across-scenarios}. The table illustrates that while the mask module contributes to performance enhancements in these settings, its impact is less pronounced compared to UDA scenarios. On average, the mask module results in an improvement of 2.81\% for EM and 2.87\% for F1 in UDA scenarios. Conversely, for supervised learning scenarios, the increments are more modest, at 0.65\% for EM and 0.61\% for F1. This is in line with our expectations. The primary role of the mask module is to preserve domain-invariant features and enhance the learning process when ground truth labels for the target domain are absent. Consequently, in settings where labeled target data is already available, the marginal benefits conferred by the mask module are reduced.

\subsection{Sensitivity Analysis for Available Target Domain Sample Numbers}
\label{sec:data_size}

In Table \ref{tab:main-results}, we showed that with $n$ ($n=1,000$) unlabeled target domain samples, we could adapt a QA model to the target domain effectively. In this section, we discuss the effect of the value of $n$ on the QA performance of the target domain. The results are shown in Figure \ref{fig:data_size_fig}. 

As depicted in the figure, the performance of MDAQA and baselines keeps improving as the size of available target domain samples increases, although the target domain samples are unlabeled.  This is intuitive, as more target domain samples can make the measurement of domain shifts more accurate, thus training better domain adaptation models.

Our MDAQA outperforms baselines with a clear margin across all data sizes. This further demonstrates that the proposed masked self-training strategy can help PLMs better capture domain differences regardless of the data size, and the improvements produced by our method are stable and reliable.

Moreover, even when the available target domain samples are extremely limited ($n=10$), MDAQA still achieves decent performance improvements compared with baselines. As depicted in Table \ref{tab:main-results}. When the source model is used to directly predict answers to HotpotQA questions, without any domain adaptation techniques, it achieves an EM of 47.94, an F1 of 65.71, while MDAQA achieves an EM of about 50, an F1 of about 67.4, according to Figure \ref{fig:data_size_fig}. When the source model is used to directly predict answers to NQ questions, it achieves an EM of 46.39, an F1 of 60.17, while MDAQA achieves an EM of about 51.9, an F1 of about 65.1. This improvement is very impressive considering that so few training examples are used.

It is worth noting that we use logarithmic scaling for the x-axis of Figure \ref{fig:data_size_fig} for better demonstration. Although the performance seems to be increasing steadily in the figure, in reality, the performance of the algorithms varies more drastically when the available data size is small. Basically, most algorithms get similar performance improvements when $n$ increases from 10 to 100 as when $n$ increases from 100 to 1,000. As depicted in Figure \ref{fig:data_size_fig}, for HotpotQA, MDAQA achieves an EM of about 50 when $n=10$, then it achieves an EM of about 52 when $n=100$, the increase is about 2 percent. Then it achieves an EM close to 54 when $n=1,000$, this increase is also about 2 percent. The performance increase for NQ is similar, albeit more gradual. 

%Furthermore, when $n>1,000$, the performance improvement of the model on HotpotQA seems to be modest. However, when $n>1,000$, the performance improvement on NQ is still considerable. This is in accordance with our main results. As depicted in Table \ref{tab:main-results}, when no domain adaptation technique is used, the source model performs worse on NQ compared to HotpotQA. This means that the initial domain variance between NQ and the source domain is larger than that between HotpotQA and the target domain. It is natural that MDAQA needs more data for optimal performance. 

In conclusion, MDAQA yields consistent gains across all data sizes. It can be used when the available data is extremely limited, and the increase in available data further improves performance. However, the improvement is significantly slower when the data size is quite large. Therefore, there is a trade-off between the accuracy and cost of collecting data, and MDAQA can be used to obtain a decent performance in the case of a limited budget. 

\subsection{Impact of Threshold}

\begin{figure*}
    \centering
    \begin{subfigure}[b]{0.32\textwidth}
        \centering
        \includegraphics[width=\textwidth]{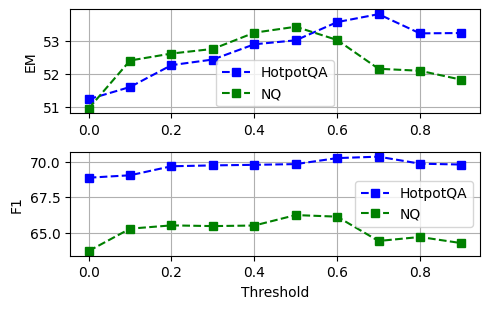}
        \caption{Performance varies with different threshold $\alpha$. (Upper: EM, lower: F1)}
        \label{fig:threshold}
    \end{subfigure}
    \hfill
    \begin{subfigure}[b]{0.32\textwidth}
        \centering
        \includegraphics[width=\textwidth]{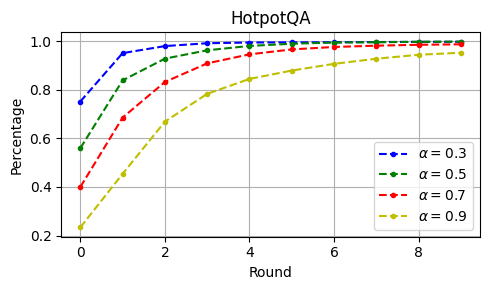}
        \caption{Numbers of pseudo-labeled samples generated in each epoch under different threshold $\alpha$ for HotpotQA.}
        \label{fig:roundHot}
    \end{subfigure}
    \hfill
    \begin{subfigure}[b]{0.32\textwidth}
        \centering
        \includegraphics[width=\textwidth]{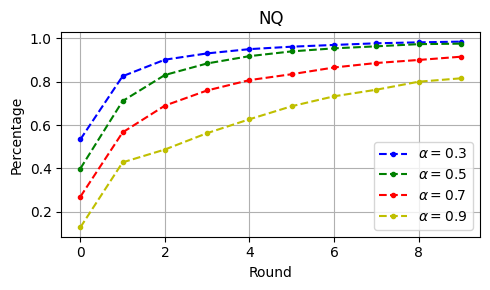}
        \caption{Numbers of pseudo-labeled samples generated in each epoch under different threshold $\alpha$ for NQ.}
        \label{fig:roundNQ}
    \end{subfigure}
       \caption{Influence of threshold $\alpha$ on adaptation performance for different datasets.}
       \label{fig:threshold-three}
\end{figure*}

\begin{figure}[ht]
    \centering
    \includegraphics[width=\columnwidth]{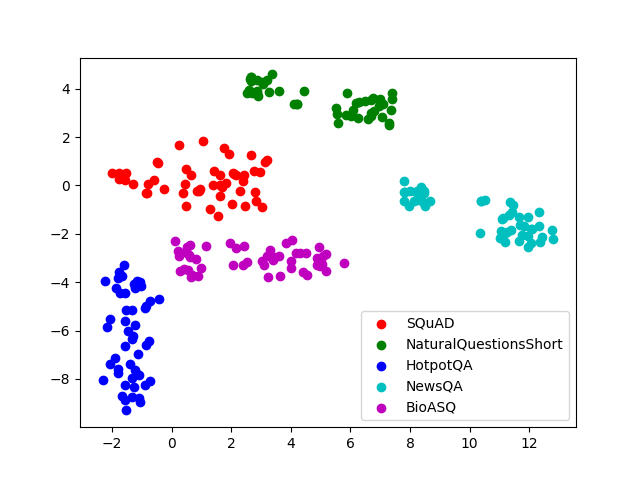}
    \caption{T-SNE visualization of question-context input embeddings.}
    \label{fig:tsne-input}
\end{figure}

\begin{figure}[ht]
    \centering
    \includegraphics[width=\columnwidth]{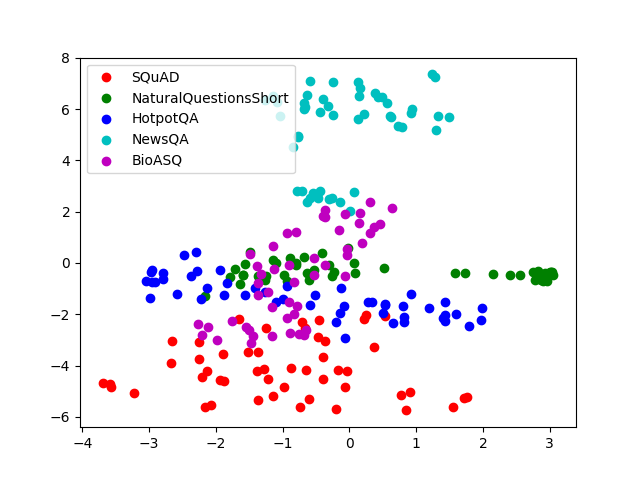}
    \caption{T-SNE visualization of question-context masked embeddings.}
    \label{fig:tsne-mask-input}
\end{figure}

Figure \ref{fig:threshold} demonstrates the performance of MDAQA varied with different threshold $\alpha$ for different target domain datasets. It seems that no matter what the target domain dataset is, the performance of MDAQA always first increases and then decreases as the value of $\alpha$ increases. This makes sense. As we can see in Figure \ref{fig:roundHot} and Figure \ref{fig:roundNQ}, when the value of $\alpha$ is too low, a large fraction of generated pseudo-labeled examples is used as training examples, even in the early rounds. This may result in too much noisy data during training, which is harmful to the final performance. On the other hand, when the value of $alpha$ is too high, the generated pseudo-labeled examples suitable as training examples are too limited. This can lead to overfitting.

As we can see in Figure \ref{fig:threshold}, MDAQA reaches the best performance on HotpotQA when $\alpha$ is around 0.7. As shown in Figure \ref{fig:roundHot}, when $\alpha=0.7$, about 40\% of the instances generated in round 0 can be used as training instances, and slightly more than 80\% of the generated instances can be used in round 2. On the other hand, the MDAQA reaches the best performance on NQ when $\alpha$ is around 0.5. As shown in Figure \ref{fig:roundNQ}, when $\alpha=0.5$, there are also about 40\% of the instances generated in round 0 which can be used as training instances, and slightly more than 80\% of the generated instances which can be used for round 2. Therefore, 40\% qualified generated samples in round 0 seem to be a good starting point for self-training. The qualified sample portion is sufficient to learn domain features while being refined enough to avoid excessive noise.

Moreover, as depicted in Figure \ref{fig:threshold}, overall, the MDAQA has better performance on HotpotQA than NQ. This suggests that the domain difference between NQ and the source domain may be larger than that between HotpotQA and the source domain. This is also validated by Figure \ref{fig:roundHot} and Figure \ref{fig:roundNQ}. For HotpotQA, even when $\alpha=0.9$, there are more than 20\% of pseudo-labeled examples are qualified as training examples in round 0. This means that the MDAQA is quite confident about its answers for at least 20\% of the questions in HotpotQA at the initial round. Furthermore, as the number of rounds increases, almost all curves in Figure \ref{fig:roundHot} reach to near 1. This indicates that MDAQA is confident in almost all of its predictions in later rounds. On the other hand, for NQ, when $\alpha=0.9$, there are only less than 10\% of pseudo-labeled examples are qualified as training examples in round 0. Even in later rounds, only slightly higher than 80\% are qualified.

In summary, the greater the domain divergence between the source and target domains, the less confident the initial model is in its predictions and the lower the number of pseudo-labeled examples eligible for training. Therefore, we need to set the value of the threshold $\alpha$ relatively low so that the proportion of qualified examples in the first round remains around 40\%.

\begin{table*}[ht]
    \centering
    \small
    \begin{tabular}{l|cc|cc|cc|cc|cc}
    \hline
    \multirow{2}{*}{Datasets} & \multicolumn{2}{c|}{SQuAD} & \multicolumn{2}{c|}{HotpotQA} & \multicolumn{2}{c|}{NQ} & \multicolumn{2}{c|}{NewsQA} & \multicolumn{2}{c}{BioASQ} \\
    \cline{2-11}
    & EM & F1 & EM & F1 & EM & F1 & EM & F1 & EM & F1 \\
    \hline
    SQuAD & - & - & \textbf{53.81} & \textbf{70.34} & \textbf{53.43} & \textbf{66.26} & \textbf{44.89} & \textbf{60.50} & \textbf{54.12} & \textbf{64.07} \\
    HotpotQA & 63.41 & 77.65 & - & - & 48.78 & 61.71 & 38.53 & 54.50 & 46.81 & 61.67 \\
    NQ & 61.30 & 77.51 & 51.82 & 68.18 & - & - & 41.71 & 58.36 & 52.06 & 64.37 \\
    NewsQA & 58.31 & 76.91 & 43.01 & 62.35 & 48.40 & 62.81 & - & - & 44.61 & 60.53 \\
    BioASQ & \textbf{65.30} & \textbf{79.27} & 52.00 & 68.87 & 50.20 & 62.53 & 37.70 & 54.24 & - & - \\
    \hline
    \end{tabular}
    \caption{{Comprehensive Performance Analysis of UDA Across Multiple QA Datasets Using Diverse Source Domains}}
    \label{tab:diverse-source-domain}
\end{table*}

\subsection{Visualization of Domain Discrepancies and the Efficacy of Mask Module}

Following the methodologies outlined by \citet{wang-etal-2019-adversarial} and \citet{zhu-hauff-2022-unsupervised}, we utilized the t-SNE visualization technique on question-context pairs. These pairs were embedded using the Roberta encoder. As shown in Figure \ref{fig:tsne-input}, the data points of HotpotQA and BioASQ are considerably close to those of SQuAD, contrasting with the data points of NewsQA which are situated at a greater distance. This observation aligns with our earlier experimental findings, indicating that NewsQA exhibits the most significant domain difference compared to SQuAD. Figure \ref{fig:tsne-mask-input} presents the embeddings post mask module application. The mask module visibly decreases the domain discrepancies between samples from different domains. This serves as further proof of the efficacy of our proposed mask module in capturing domain-invariant features.

\begin{table*}[ht]
    \centering
    \small
    \begin{tabular}{lcccccccc}
    \hline
    \multirow{2}{*}{Models} & \multicolumn{2}{c}{HotpotQA} & \multicolumn{2}{c}{NQ} & \multicolumn{2}{c}{NewsQA} & \multicolumn{2}{c}{BioASQ} \\
    \cline{2-9}
    & EM & F1 & EM & F1 & EM & F1 & EM & F1 \\
    \hline
    BERT & 43.50 & 60.58 & 42.61 & 56.31 & 37.06 & 53.63 & 45.68 & 57.01 \\
    +MDAQA & \textbf{46.50} & \textbf{63.59} & \textbf{47.10} & \textbf{61.30} & \textbf{41.10} & \textbf{55.29} & \textbf{48.71} & \textbf{60.03} \\
    \hline
    ALBERT & 46.16 & 64.22 & 39.22 & 54.15 & 40.95 & 57.69 & 45.94 & 58.23 \\
    +MDAQA & \textbf{48.00} & \textbf{66.79} & \textbf{43.51} & \textbf{58.25} & \textbf{43.71} & \textbf{58.72} & \textbf{49.23} & \textbf{58.74} \\
    \hline
    DistilRoBERTa & 43.38 & 60.27 & 41.57 & 54.78 & 39.48 & 54.95 & 45.74 & 57.05 \\
    +MDAQA & \textbf{46.80} & \textbf{62.07} & \textbf{43.80} & \textbf{56.16} & \textbf{40.40} & \textbf{55.06} & \textbf{50.90} & \textbf{59.59} \\
    \hline
    \end{tabular}
    \caption{{Enhanced performance across multiple QA datasets with diverse backbone models: A comparative analysis of BERT, ALBERT and DistilRoBERTa with and without the MDAQA framework.}}
    \label{tab:model-performance}
\end{table*}

\subsection{Feature Map Analysis}

\begin{figure}[ht]
    \centering
    \includegraphics[width=\columnwidth]{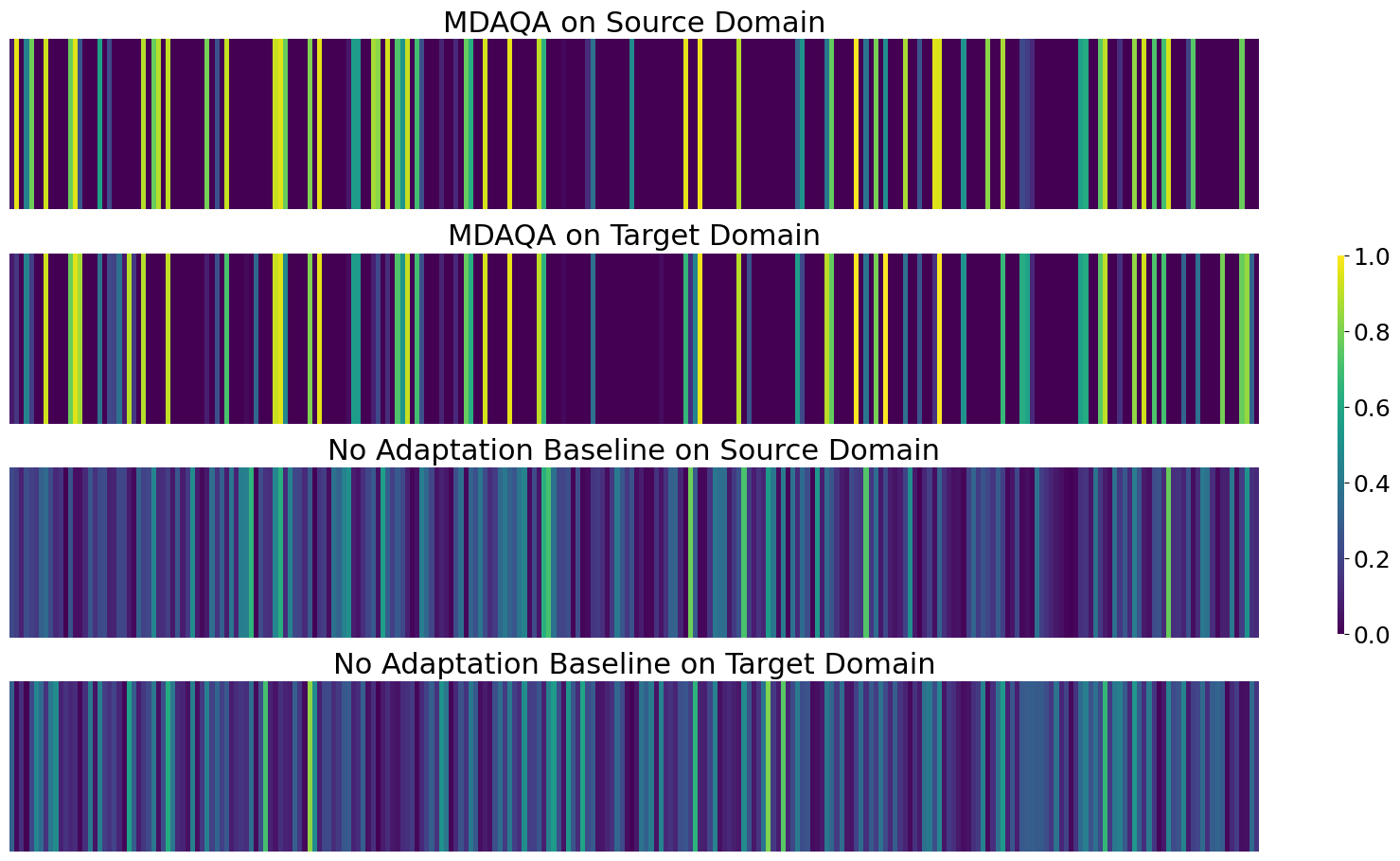}
    \caption{{Comparative feature map analysis between MDAQA and no adaptation baseline models on source and target domains. The top two heatmaps illustrate the feature embeddings for the MDAQA model on source and target domains, respectively. The bottom two heatmaps display the feature embeddings for the no adaptation baseline model on the respective domains.}}
    \label{fig:feature-maps}
\end{figure}

{
    To elucidate the impact of our proposed mask module, we visualized the feature embeddings post-encoder for MDAQA in comparison with the No Adaptation baseline. As delineated in Figure \ref{fig:feature-maps}, the MDAQA model, endowed with a mask module, exhibits stark and coherent feature extractions in both the source and target domains. There exist zones of marked consistency within the feature maps, indicative of the domain-invariant features retained from the source domain. Concurrently, we observe discrete regions wherein the feature maps diverge across domains, suggesting the assimilation of domain-specific features by the model. By contrast, the feature maps of the No Adaptation baseline, manifest no discernible extraction patterns. This contrast accentuates the proficiency of MDAQA in not only conserving pivotal domain knowledge from the source domain but also in its adeptness at assimilating novel, domain-specific features pertinent to the target domain. The absence of this dual capability in the baseline model serves to emphasize the superior domain adaptation prowess inherent in our MDAQA methodology.
}

\subsection{Diverse Source Domain Analysis}

\begin{table*}[ht]
    \centering
    \small
    \begin{tabular}{l|cc|cc|cc|cc|cc}
    \hline
    \multirow{2}{*}{Methods} & \multicolumn{2}{c|}{HotpotQA} & \multicolumn{2}{c|}{NQ} & \multicolumn{2}{c|}{NewsQA} & \multicolumn{2}{c|}{BioASQ} & \multicolumn{2}{c}{MIMIC-III} \\
    \cline{2-11}
    & EM & F1 & EM & F1 & EM & F1 & EM & F1 & EM & F1 \\
    \hline
    No Adaptation & 47.94 & 65.71 & 46.39 & 60.17 & 41.12 & 57.20 & 49.77 & 61.69 & 14.32 & 47.44 \\
    UDARC & 47.98 & 66.01 & 46.53 & 61.32 & 41.08 & 56.97 & 50.21 & 62.98 & 15.11 & 49.02 \\
    AdaMRC & 48.69 & 64.02 & 47.99 & 62.51 & 40.02 & 56.48 & 51.87 & 63.99 & 16.31 & 49.67 \\
    CAQA & 50.62 & 64.39 & 50.21 & 67.13 & 40.98 & 57.65 & 53.22 & 64.18 & 17.81 & 51.36 \\
    \textbf{MDAQA} & \textbf{52.69} & \textbf{66.05} & \textbf{53.11} & \textbf{70.32} & \textbf{42.19} & \textbf{58.58} & \textbf{54.30} & \textbf{66.98} & \textbf{18.42} & \textbf{52.48} \\
    \hline
    \end{tabular}
    \caption{{Experiment results on comparing question-answering performance while performing domain adaptation from SQuAD to MRQA datasets and MIMIC-III with generated questions.}}
    \label{tab:generated-questions}
\end{table*}

{
    In the pursuit of understanding the impact of source domain diversity on the efficacy of domain adaptation in QA systems, we conducted an exhaustive analysis across all domain datasets. The results, presented in Table \ref{tab:diverse-source-domain}, illustrate the domain adaptation performance when different datasets are used as the source domain. Notably, SQuAD, known for its rich and diverse question set, emerged as an exemplary source domain, yielding substantial performance gains across all targeted domains. This can be attributed to the general-purpose nature of SQuAD, encompassing a wide array of topics and question formats, which provides a robust and comprehensive foundation for the model. 
}

\subsection{Comparative Evaluation of Different Backbone Models}

{
    To rigorously evaluate the robustness of the MDAQA framework, we conducted extensive experiments using a range of backbone models, including BERT \citep{devlin2018bert}, ALBERT \citep{lan2019albert} and DistilRoBERTa. The results, as systematically presented in Table \ref{tab:model-performance}, demonstrate a consistent improvement across all models, emphatically affirming the robustness and versatility of the MDAQA framework. This enhancement in performance is observed irrespective of the specific backbone model utilized, underscoring the effectiveness of MDAQA in augmenting model capabilities. 
}

\subsection{No-Question Adaptation}

We further evaluated our proposed method under a more stringent condition where questions from the target domain are unavailable. To achieve UDA in this context, we employed the T5 model \citep{roberts2019exploring} as a supplementary question generation tool, training it on context-question pairs from the source domain. This model subsequently generated pseudo-questions for the target domains. {These generated questions were used instead of actual target domain questions, adhering to the standard training protocol for MDAQA and the introduced baseline models.  Employing this strategy facilitated a direct comparison of the effectiveness of our method against traditional adaptation techniques when real target-domain queries are absent. The results, detailed in Table \ref{tab:generated-questions}, affirm the robustness of our approach in utilizing synthetic questions to mitigate the domain discrepancy. }

\section{Conclusion}

In this paper, we explore the possibility of transferring knowledge for UDA on QA tasks, without access to initial domain data. We proposed a novel self-training-based approach, MDAQA. We specially design an attention mask module to automatically retain key knowledge from the source domain and learn to mitigate domain shift between source and target domains. The module can be easily integrated into existing language models. Our comprehensive experiments on well-known benchmark datasets demonstrate that MDAQA outperforms previous methods by a clear margin. It also performs well even when the available target domain data is highly limited, suggesting that MDAQA has wide applicability across a range of scenarios.

Our current study primarily focuses on extractive QA tasks, aligned to the approach of previous research in the field \citep{wang-etal-2019-adversarial,nishida2019unsupervised,yue-etal-2021-contrastive}. We chose the Roberta model specifically for its exceptional aptitude for such tasks. Furthermore, we recognize the potential and need to broaden the scope of our work towards more versatile settings, such as open-domain and generative QA. To this end, our future plans include transitioning from Roberta to a T5-based model. Known for its ability to handle a wide variety of tasks, the T5 model will provide us the flexibility and breadth required to venture into broader and more challenging realms of QA.

\section*{Acknowledgments}

The authors extend their sincere appreciation to Kristina Toutanova and Michael Collins, the action editors, as well as to the anonymous reviewers, for their invaluable feedback. Their meticulous guidance played a pivotal role in refining the quality and comprehensiveness of the research discussed in this manuscript. Additionally, this work received support from the Natural Sciences and Engineering Research Council of Canada (NSERC), Discovery Grants program.

\bibliography{anthology,custom}
\bibliographystyle{acl_natbib}

\end{document}